\pdfoutput=1

\documentclass[11pt]{article}

\usepackage[final]{acl}

\usepackage{times}
\usepackage{latexsym}

\usepackage[T1]{fontenc}

\usepackage[utf8]{inputenc}

\usepackage{microtype}

\usepackage{inconsolata}

\usepackage{graphicx}

\usepackage{hyperref}
\usepackage{natbib}
\usepackage{lineno}
\usepackage{booktabs}
\usepackage{xcolor}

\definecolor{ggreen}{HTML}{00A64F}
\definecolor{light-gray}{gray}{0.9}

\newcommand*{\tightcolorbox}[2]{%
    \begingroup\setlength{\fboxsep}{1pt}%
        \colorbox{#1}{{\hspace*{2pt}\vphantom{Ay}#2\hspace*{2pt}}}%
    \endgroup
}
\newcommand*{\code}[1]{\tightcolorbox{light-gray}{\texttt{#1}}}

\newcommand*{\modelname}[1]{{\textsc{#1}}}
\newcommand*{\datasetname}[1]{{\textsc{#1}}}

\newcommand\blfootnote[1]{%
  \begingroup
  \renewcommand\thefootnote{}%
  \NoHyper\footnote{#1}\endNoHyper%
  \addtocounter{footnote}{-1}%
  \endgroup
}

\title{From Language to Cognition:\\How LLMs Outgrow the Human Language Network}
 
\author{
{Badr AlKhamissi$^1$} 
\quad
{Greta Tuckute$^2$} 
\quad
{Yingtian Tang$^1$} 
\quad
{Taha Binhuraib$^3$} 
\vspace{0.25cm}
\\
{\textbf{Antoine Bosselut}$^{*,1}$}
\quad
{\textbf{Martin Schrimpf}$^{*,1}$}
\vspace{0.25cm}
\\
$^1$EPFL \quad $^2$MIT \quad $^3$Georgia Institute of Technology
}

\begin{document}
\maketitle

\blfootnote{$^*$ Equal Supervision}
\begin{abstract}
    Large language models (LLMs) exhibit remarkable similarity to neural activity in the human language network.
    However, the key properties of language underlying this alignment---and how brain-like representations emerge and change across training---remain unclear.
    We here benchmark 34 training checkpoints spanning 300B tokens across 8 different model sizes to analyze how brain alignment relates to linguistic competence. Specifically, we find that brain alignment tracks the development of formal linguistic competence---i.e., knowledge of linguistic rules---more closely than functional linguistic competence. While functional competence, which involves world knowledge and reasoning, continues to develop throughout training, its relationship with brain alignment is weaker, suggesting that the human language network primarily encodes formal linguistic structure rather than broader cognitive functions.
    Notably, we find that the correlation between next-word prediction, behavioral alignment, and brain alignment fades once models surpass human language proficiency.
    We further show that model size is not a reliable predictor of brain alignment when controlling for the number of features. 
    Finally, using the largest set of rigorous neural language benchmarks to date, we show that language brain alignment benchmarks remain unsaturated, highlighting opportunities for improving future models.
    Taken together, our findings suggest that the human language network is best modeled by formal, rather than functional, aspects of language.\footnote{Project Page: \href{https://language-to-cognition.epfl.ch}{language-to-cognition.epfl.ch}}
\end{abstract}

\section{Introduction}

\begin{figure}[t!]
    \centering
    \includegraphics[width=1\linewidth]{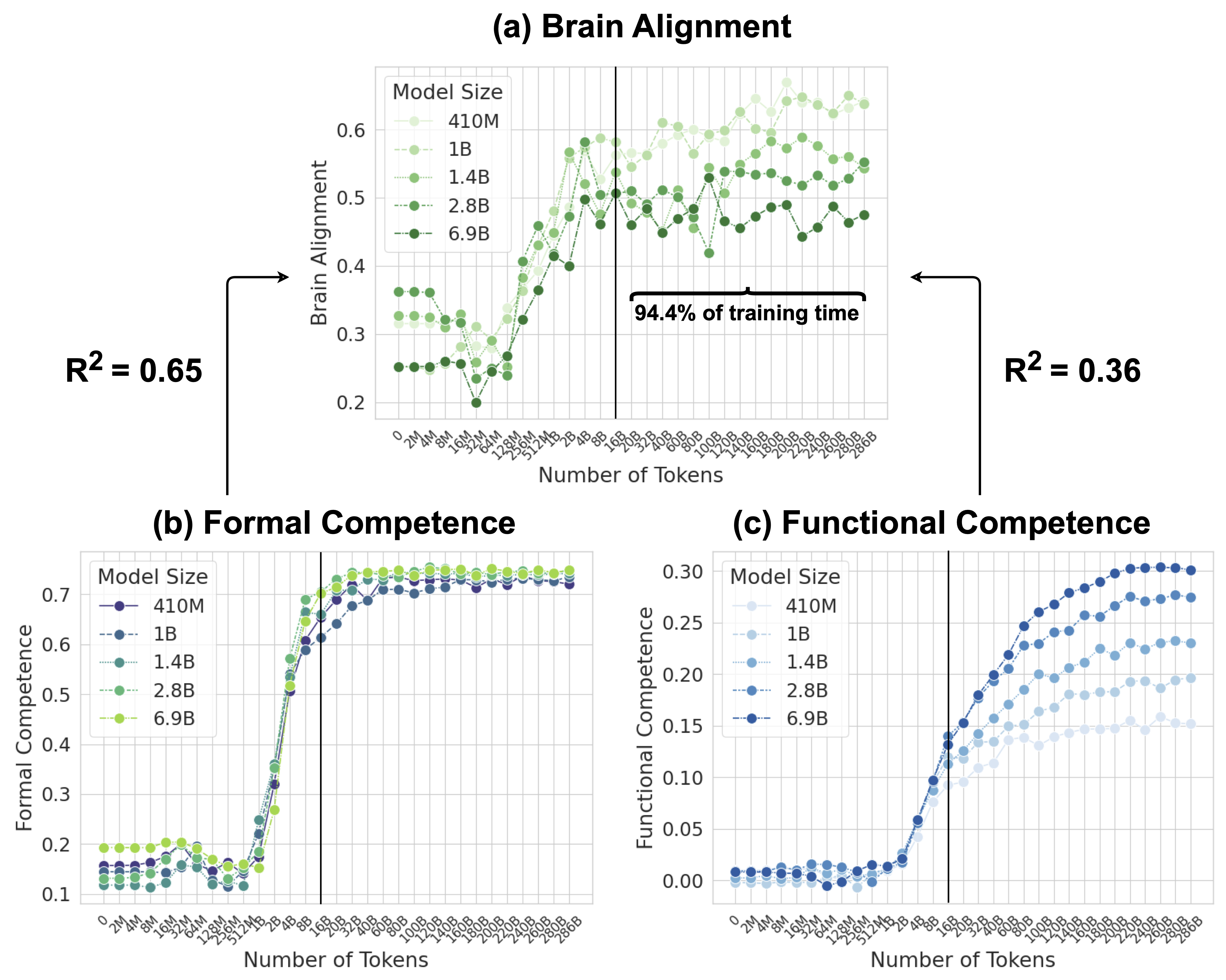}
    \caption{
    \textbf{Model Alignment with the Human Language Network is Primarily Driven by Formal than Functional Linguistic Competence.}  
    \textbf{(a)} Average brain alignment across five Pythia models and five brain recording datasets, normalized by cross-subject consistency, throughout training.  
    \textbf{(b)} Average normalized accuracy of the same models on formal linguistic competence benchmarks (two benchmarks).  
    \textbf{(c)} Average normalized accuracy on functional linguistic competence benchmarks (six benchmarks).  
    The x-axis is logarithmically spaced up to 16B tokens, capturing early training dynamics, and then evenly spaced every 20B tokens from 20B to \textasciitilde300B tokens. 
    }
    \label{fig:pretrained}
\end{figure}

Deciphering the brain's algorithms underlying our ability to process language and communicate is a core goal in neuroscience. Human language processing is supported by the brain's language network (LN), a set of left-lateralized fronto-temporal regions in the brain \citep{Binder1997,Bates2003,GornoTempini2004,Price2010,Fedorenko2014,Hagoort2019} that respond robustly and selectively to linguistic input \citep{Fedorenko2024}. Driven by recent advances in machine learning, large language models (LLMs) trained via next-word prediction on large corpora of text are now a particularly promising model family to capture the internal processes of the LN. In particular, when these models are exposed to the same linguistic stimuli (e.g., sentences or narratives) as human participants during neuroimaging and electrophysiology experiments, they account for a substantial portion of neural response variance \citep{schrimpf-pnas, caucheteux2022brains, goldstein_shared_2022, pasquiou2022neural, aw2023instructiontuning,tuckute2024language, AlKhamissi2024TheLN, rathimehrer2025TopoLMBrainlikeSpatiofunctional}. 

\subsection{Key Questions and Contributions}
This work investigates four key questions, all aimed at distilling \textit{why} LLM aligns to brain responses. Specifically, we investigate the full model development cycle as a combination of model architecture (structural priors) and how linguistic competence emerges across training (developmental experience). We ask:
(1) What drives brain alignment in untrained models? 
(2) Is brain alignment primarily linked to formal or functional linguistic competence \citep{mahowald2024dissociating}?
(3) Do language models diverge from humans as they surpass human-level prediction?
(4) Do current LLMs fully account for the explained variance in brain alignment benchmarks?
To answer these questions, we introduce a rigorous brain-scoring framework to conduct a controlled and large-scale analysis of LLM brain alignment.

Our findings reveal that the initial brain alignment of models with untrained parameters is driven by context integration.
During training, alignment primarily correlates with formal linguistic competence—tasks that probe mastery of grammar, syntax, and compositional rules, such as identifying subject–verb agreement, parsing nested syntactic structures, or completing well-formed sentences. This competence saturates relatively early in training ($\sim 4$B tokens), consistent with a plateauing of model-to-brain alignment.
Functional linguistic competence, in contrast, concerns how language is used in context to convey meaning, intent, and social/pragmatic content—for example, tasks involving discourse coherence, reference resolution, inference about speaker meaning, or interpreting figurative language. Functional competence emerges later in training, tracks brain alignment less strongly, and continues to grow even after alignment with the language network has saturated.

This disconnect later in training is further exemplified by a fading of the correlation between models' brain alignment and their next-word-prediction performance, as well as their behavioral alignment.
Further, we show that model size is not a reliable predictor of brain alignment when controlling for the number of features, challenging the assumption that larger models necessarily resemble the brain more. Finally, we demonstrate that current brain alignment benchmarks remain unsaturated, indicating that LLMs can still be improved to model human language processing.


\section{Preliminaries \& Related Work}
\label{sec:preliminaries}

\paragraph{A Primer on Language in the Human Brain}
The human language network (LN) is a set of left-lateralized frontal and temporal brain regions supporting language.  These regions are functionally defined by contrasting responses to language inputs over perceptually matched controls (e.g., lists of non-words) \citep{Fedorenko2010NewMF}. The language network exhibits remarkable selectivity for language processing compared to various non-linguistic inputs and tasks, such as music perception \citep{Fedorenko2012music, chen2023human} or arithmetic computation \citep{fedorenko2011functional,monti2012thought} (for review, see \citet{Fedorenko2024}) and the language network only shows weak responses when participants comprehend or articulate meaningless non-words \citep{Fedorenko2010NewMF,hu2023precision}. This selectivity profile is supported by extensive neuroimaging research and further corroborated by behavioral evidence from aphasia studies: when brain damage is confined to language areas, individuals lose their linguistic abilities while retaining other skills, such as mathematics \citep{Benn2013, Varley2005}, general reasoning \citep{Varley2000}, and theory of mind \citep{Siegal2006}.

\paragraph{Model-to-Brain Alignment}
Prior work has shown that the internal representations of certain artificial neural networks resemble those in the brain. This alignment was initially observed in the domain of vision \citep{yamins2014performance, Khaligh-Razavi2014, cichy2016comparison, schrimpf_brain-score_2018, schrimpf2020, cadena2019deep, Kubilius2019, Zhuang2021} and has more recently been extended to auditory processing \citep{kell2018task,Tuckute23many,koumura2023human} and language processing \citep{schrimpf-pnas, caucheteux2022brains, goldstein_shared_2022, Kauf23lexical, hosseini2024artificial, aw2023instructiontuning, AlKhamissi2024TheLN, tuckute2024driving, rathimehrer2025TopoLMBrainlikeSpatiofunctional}. 

\paragraph{Untrained Models}
Recent work in vision neuroscience has shown that untrained convolutional networks can yield high brain alignment to recordings in the visual ventral stream without the need for training \citep{Geiger2022,Kazemian2024}. Other works have investigated the inductive biases in different architectures and initializations in models of visual processing \citep{cichy2016comparison, cadena2019deep, Geiger2022}, speech perception \citep{Millet2021InductiveBP, Tuckute23many}, and language \citep{schrimpf-pnas, pasquiou2022neural,hosseini2024artificial}, highlighting that randomly initialized networks are not random functions \citep{teney2024neural}. 

\section{Methods}
\label{sec:benchmarks}

\subsection{Benchmarks for Brain Alignment}
\label{sec:brain-benchmarks}

\paragraph{Neuroimaging \& Behavioral Datasets}
The neuroimaging datasets used in this work can be categorized along three dimensions: the imaging modality, the context length of the experimental materials, and the modality through which the language stimulus was presented to human participants (auditory or visual). Table \ref{tab:dataset} in Appendix \ref{app:datasets} provides an overview of all datasets in this study.  
To focus specifically on language, we consider neural units (electrodes, voxels, or regions) associated with the brain's language network, as localized by the original dataset authors using the method described in the Section \ref{sec:localization} and implemented in Brain-Score \cite{schrimpf2020,schrimpf-pnas} (however, see Appendix \ref{app:other-brain-regions} for control brain regions).
An exception is the \datasetname{Narratives} dataset, which lacks functional localization. We here approximate the language regions using a probabilistic atlas of the human language network \citep{Lipkin2022}, extracting the top-10\% language-selective voxels (from the probabilistic atlas) within anatomically defined language parcels, in line with the functional localization procedure used in the other datasets.
In an additional analysis, we investigate model alignment with language behavior using the \cite{futrell_natural_2018} dataset, which contains self-paced, per-word human reading times. See Appendix \ref{app:datasets} for details of each dataset.
To the best of our knowledge, this study examines the largest number of benchmarks compared to previous work, providing a more comprehensive and reliable foundation for identifying the properties that drive brain alignment in LLMs. The diversity of datasets ensures that our conclusions generalize beyond specific experimental stimuli and paradigms.


\paragraph{Brain-Alignment Metrics}
Following standard practice in measuring brain alignment, we train a ridge regression model to predict brain activity from model representations, using the same linguistic stimuli presented to human participants in neuroimaging studies \citep{schrimpf2020,schrimpf-pnas}. We then measure the Pearson correlation between the predicted brain activations and the actual brain activations of human participants on a held-out set that covers entirely different stories or topics (see Section \ref{sec:rigorous-brainscoring}). This process is repeated over $k$ cross-validation splits, and we report the average (mean) Pearson correlation as our final result. We refer to this metric as Linear Predictivity. In Section \ref{sec:untrained}, we demonstrate why other metrics such as Centered Kernel Alignment (CKA; \citealp{cka}) and Representational Similarity Analysis (RSA; \citealp{rdm}) are not suitable measures for brain alignment on current language datasets.

\paragraph{Estimation of Cross-Subject Consistency}
To assess the reliability of our datasets and account for the inherent noise in brain recordings, we compute a cross-subject consistency score \citep{Feather2025BrainModelEN}, also referred to as the noise ceiling \citep{schrimpf-pnas}. The consistency score is estimated by predicting the brain activity of a held-out subject using data from all other subjects, through 10-fold cross-validation of all subjects.
To obtain a conservative ceiling estimate, we extrapolate subject pool sizes and report the final value based on extrapolation to infinitely many subjects. 
For \datasetname{Tuckute2024} we use the theoretical estimate provided by \citep{tuckute2024driving}.
Consistency scores are provided in Appendix \ref{app:consistency}.
To aggregate scores across benchmarks, we normalize each model’s Pearson correlation ($r$) score for Linear Predictivity by the cross-subject consistency estimate, using the formula: ($\textnormal{normalized score} = \frac{\textnormal{raw score}}{\textnormal{consistency}}$).
The final alignment score for each model is reported as the average across all benchmarks. Otherwise, when reporting raw alignment, we compute the mean Pearson correlation across datasets without normalization.

\subsection{Functional Localization}
\label{sec:localization}

The human language network (LN) is defined \emph{functionally} which means that units are chosen according to a `localizer' experiment \citep{saxe2006divide}. Specifically, the LN is the set of neural units (e.g., voxels/electrodes) that are more selective to sentences over a perceptually-matched control condition \citep{Fedorenko2010NewMF}.
When selecting units from artificial models for comparison against LN units, previous work selected output units from an entire Transformer block based on brain alignment scores \citep{schrimpf-pnas}. However, LLMs learn diverse concepts and behaviors during their considerable pretraining, not all of which are necessarily related to language processing, e.g., storage of knowledge \citep{AlKhamissi2022ARO} and the ability to perform complex reasoning \citep{huang-chang-2023-towards}.
Therefore, we here follow the method proposed by \cite{AlKhamissi2024TheLN} that identifies language units in LLMs using functional localization as is already standard in neuroscience. This approach offers a key advantage: it enables direct comparisons across models by selecting a fixed set of units, identified through the independent localizer experiment. In this work, we localize $128$ units for all models unless otherwise specified, and we show in Appendix \ref{app:number-of-units} that the results hold when selecting a different number of units.

\begin{figure*}[]
    \centering
    \includegraphics[width=1\linewidth]{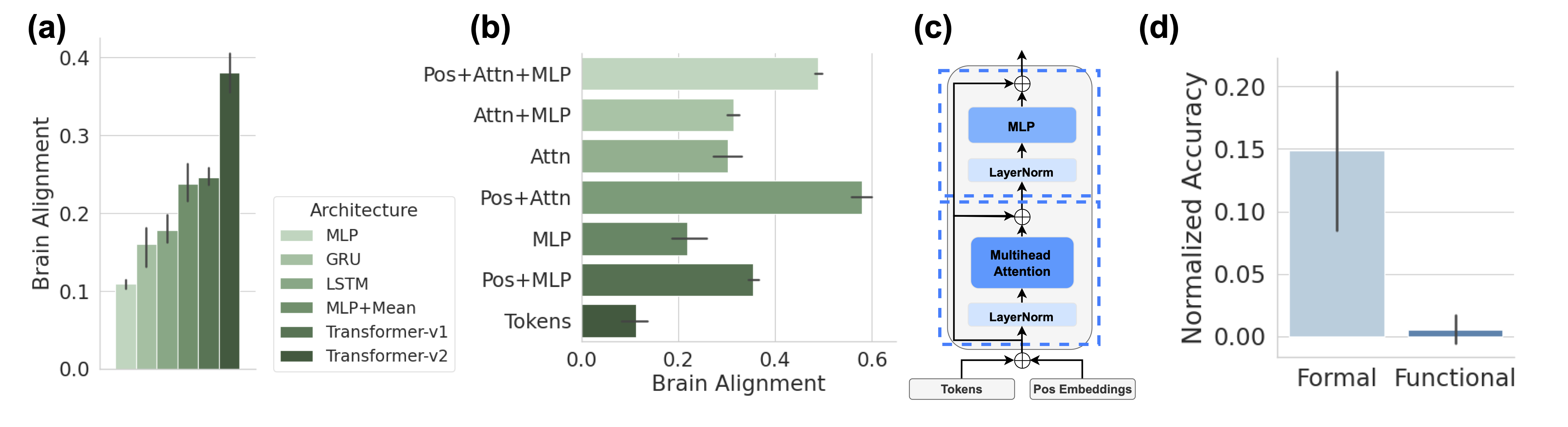}
    \caption{
        \textbf{Context Integration drives Brain Alignment of Untrained Models}. 
        \textbf{(a)} Sequence-based models (GRU, LSTM, Transformers, and mean pooling) achieve higher brain alignment than models that rely solely on the last token representation (Linear, MLP), highlighting the importance of temporal integration. Error bars report five random initializations in all subplots.
        \textbf{(b)} Ablation study of architectural components in a single untrained \modelname{Transformer-v2} block, demonstrating that attention mechanisms combined with positional encoding yield the highest brain alignment. 
        \textbf{(c)} Diagram of the Transformer block architecture used in (b), with components grouped into attention (lower box) and MLP (upper box). 
        \textbf{(d)} The average performance of five Pythia models with untrained parameters on formal and functional linguistic competence benchmarks, showing that formal competence exceeds chance level even in untrained parameter models.
    }
        
    \label{fig:untrained-models}
\end{figure*}

\subsection{Benchmarks for Linguistic Competence}

There is substantial evidence in neuroscience research that formal and functional linguistic competence are governed by distinct neural mechanisms \cite{mahowald2024dissociating, Fedorenko2024, Fedorenko2024lang}. Formal linguistic competence pertains to the knowledge of linguistic rules and patterns, while functional linguistic competence involves using language to interpret and interact with the world. Therefore, to accurately track the evolution of each type of competence during training, we focus on benchmarks that specifically target these cognitive capacities in LLMs. 

\paragraph{Formal Linguistic Competence}
To assess formal linguistic competence, we use two benchmarks: \datasetname{BLiMP} \citep{Warstadt2019BLiMPTB} and \datasetname{SyntaxGym} \citep{gauthier-etal-2020-syntaxgym}. \datasetname{BLiMP} evaluates key grammatical phenomena in English through 67 tasks, each containing 1,000 minimal pairs designed to test specific contrasts in syntax, morphology, and semantics. Complementing this, \datasetname{SyntaxGym} consists of 31 tasks that systematically measure the syntactic knowledge of language models. 
Together, these benchmarks provide a robust framework for evaluating how well LLMs acquire and apply linguistic rules.

\paragraph{Functional Linguistic Competence}
Functional competence extends beyond linguistic rules, engaging a broader set of cognitive mechanisms. To assess this, we use six benchmarks covering world knowledge (\datasetname{ARC-Easy}, \datasetname{ARC-Challenge} \citep{Clark2018ThinkYH}), social reasoning (\datasetname{Social IQa} \citep{sap-etal-2019-social}), physical reasoning (\datasetname{PIQA} \citep{Bisk2019PIQARA}), and commonsense reasoning (\datasetname{WinoGrande} \citep{Sakaguchi2019WinoGrande}, \datasetname{HellaSwag} \citep{Zellers2019HellaSwagCA}). 
Together, these benchmarks provide a comprehensive evaluation of an LLM’s ability to reason, infer implicit knowledge, and navigate real-world contexts.

\paragraph{Metrics}
Inline with prior work, we evaluate all benchmarks in a zero-shot setting, using surprisal as the evaluation metric. where the model's prediction is determined by selecting the most probable candidate, as packaged in the language model evaluation harness \citep{eval-harness}.
We report accuracy normalized by chance performance, where 0\% indicates performance at the random chance level.

\paragraph{Benchmark for Language Modeling}
We use a subset of \datasetname{FineWebEdu} \cite{fineweb} to evaluate the perplexity of the models on a held-out set. Specifically, use a maximum sequence length of 2048, and evaluate on the first 1000 documents of the \code{CC-MAIN-2024-10} subset.  

\subsection{Large Language Models (LLMs)}
Throughout this work, we use eight models from the Pythia model suite \citep{pythia}, spanning a range of sizes: \{14M, 70M, 160M, 410M, 1B, 1.4B, 2.8B, 6.9B\}. Each model is evaluated across 34 training checkpoints, spanning approximately 300B tokens. These checkpoints include the untrained model, the final trained model, and 16 intermediate checkpoints that are logarithmically spaced up to 128B tokens. The remaining 14 checkpoints are evenly spaced every 20B tokens from 20B to 280B tokens, ensuring a comprehensive analysis of alignment trends throughout training.
Since smaller models fail to surpass chance performance on many functional benchmarks, we exclude {14M, 70M, 160M} from analyses that compare brain alignment with functional performance.

\section{Rigorous Brain-Scoring}
\label{sec:rigorous-brainscoring}

While substantial progress has been made in measuring alignment between LLM representations and neural activity, there’s no standard for comparing brain alignment across datasets and conditions. Therefore, to ensure we perform meaningful inferences, we propose two criteria: (1) alignment should reflect stimulus-driven responses, dropping for random token sequences; and (2) models should generalize to new linguistic contexts. We justify our metrics and cross-validation choices accordingly. For all benchmarks, we identify language-selective units to ensure fair model comparisons, consistent with neural site selection in neuroscience \cite{AlKhamissi2024TheLN}.

\subsection{Robust Metrics and Generalization Tests}
\label{sec:stimulus-driven-responses}

\paragraph{Measuring Stimulus-Driven Responses}
We first ask if the alignment procedure is meaningful, i.e., whether the encoding models capture meaningful linguistic information and generalize to new linguistic contexts. Figure \ref{fig:metrics-context}(a) in Appendix \ref{app:rigorous-brain-scoring} shows average brain alignment across all brain datasets under three conditions: (1) a pretrained model processing original stimuli, (2) a pretrained model processing random token sequences, and (3) an untrained model processing original stimuli.
To evaluate metric reliability, we expect random sequences to yield significantly lower alignment than real stimuli. However, CKA fails this criterion, assigning similar alignment scores to both, and even untrained models surpass pretrained ones. In contrast, linear predictivity differentiates between real and random stimuli, more so than RSA.

\paragraph{Generalization and Contextualization}
The second criterion we propose is that LLMs with high brain alignment should be able to generalize to held-out stimuli, with a preference for generalizing far outside the stimuli used for mapping the model to brain activity. 
A key factor in designing a corresponding cross-validation scheme is contextualization---how the data is split into train and test sets \cite{feghhi2024large}. 
The \datasetname{Pereira2018} dataset consists of 24 topics composed of multi-sentence passages, and sentences are presented in their original order to both humans and models. A random sentence split (contextualization) allows sentences from the same topic in both train and test sets, and is thus less demanding of generalization. A stronger generalization test ensures entire topics are held out, preventing models from leveraging shared context.
Figure \ref{fig:metrics-context}(b) shows that contextualization makes it easier for the model to predict brain activity.
In contrast, topic-based splits halve the raw alignment score for pretrained models. The score of untrained models is reduced even more strongly when enforcing generalization across topics, suggesting that much of their alignment is context-dependent. Nonetheless, untrained models retain significant alignment -- about 50\% of pretrained models -- even with strong generalization requirements. 

\begin{figure*}[t]
    \centering
    \includegraphics[width=1\linewidth]{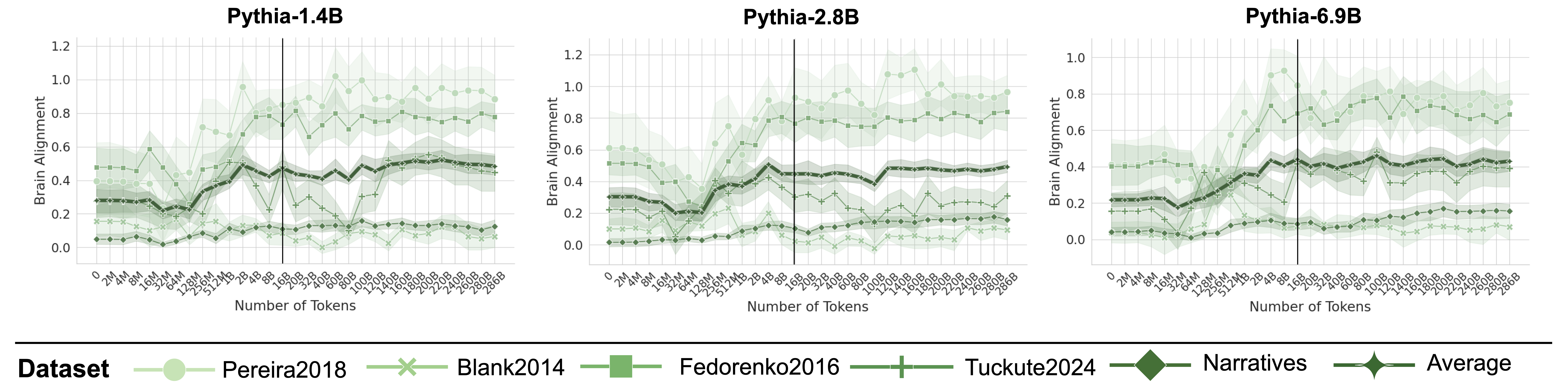}
    \caption{
        \textbf{Brain Alignment Saturates Early on in Training.} 
        Plots indicate the brain alignment scores of three models from the Pythia model suite with varying sizes (log x-axis up to 16B tokens, uneven spacing after black line). Scores are normalized by their cross-subject consistency scores. Alignment quickly peaks around 2–8B tokens before saturating or declining, regardless of model size (see Appendix \ref{app:brain-alignment-training} and \ref{app:smollm2-results} for more models).
    }
    \label{fig:bscore-training}
\end{figure*}

\section{Results}

The following sections progressively unpack the emergence and limits of brain alignment with the human language network in LLMs. Section~\ref{sec:untrained} establishes the foundation by showing that untrained models already exhibit modest brain alignment, pointing to the role of architectural priors. Building on this, Section~\ref{sec:alignment-training} tracks how alignment evolves with training and reveals that it strongly correlates with the early acquisition of formal linguistic competence, but less so with functional abilities. Section~\ref{sec:llm-behavioral-alignment} then shows that as models exceed human-level performance in next-word prediction, their brain and behavioral alignment begins to diverge, suggesting that at this point, LLMs outgrow their initial alignment with human language processing.

\subsection{Brain Alignment of Untrained Models}
\label{sec:untrained}

In Figure \ref{fig:metrics-context} we show that untrained models, despite achieving lower alignment scores than their pretrained counterparts ($\sim 50\%$), still achieve relatively decent alignment and surpass that of the models evaluated with a random sequence of tokens. Therefore, we here ask, what are the main drivers for this surprising alignment.

\paragraph{Inductive Biases of Untrained Models}
We evaluate the brain alignment of various LLMs with untrained parameters to determine which architecture exhibits the strongest inductive bias toward the human language network. Figure \ref{fig:untrained-models}(a) presents the average alignment across five different random initializations for six different untrained models. Each model consists of a stack of two building blocks from its respective architecture, with a hidden state of $1024$. To ensure a fair comparison, we apply the localizer to the output representations of the last token in the sequence from these two blocks, extracting 128 units to predict brain activity.
Our findings reveal two key insights. First, sequence-based models—such as \modelname{GRU}, \modelname{LSTM}, \modelname{Transformers}, and even a simple mean operation over token representations—exhibit higher brain alignment than models that rely solely on the last token's representation, such as \modelname{Linear} or \modelname{MLP}. In other words, context or temporal integration is a crucial factor in achieving high alignment. Second, we observe a notable difference between \modelname{Transformer-v1} and \modelname{Transformer-v2}. While \modelname{Transformer-v2} applies static positional embeddings by directly adding them to token embeddings, \modelname{Transformer-v1} uses rotary position encoding. Our results suggest that static positional encoding enables models to capture intrinsic temporal dynamics in sentences---possibly tracking evolving word positions---providing further evidence that temporal integration is critical for brain-like language representations.

\begin{figure*}[t]
    \centering
    \includegraphics[width=1\linewidth]{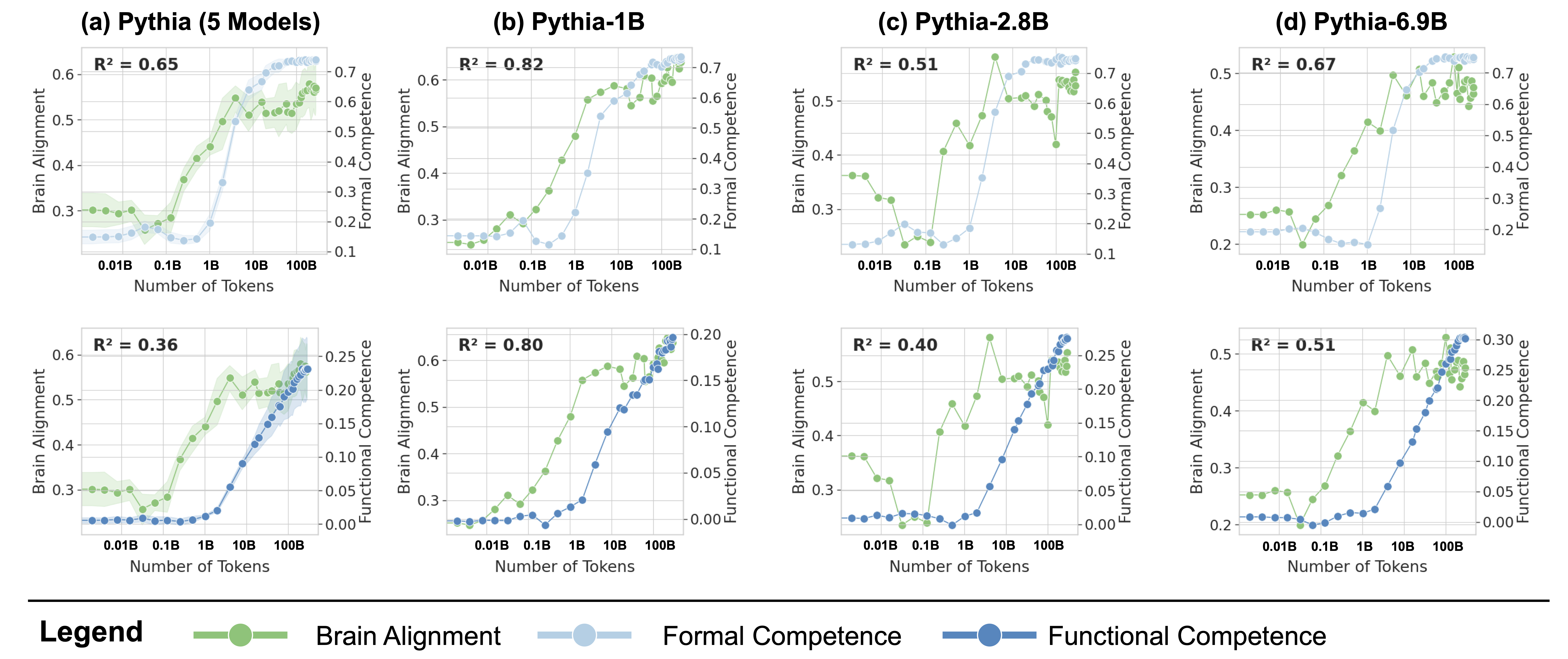}
    \caption{
        \textbf{Formal Competence Tracks Brain Alignment More Closely Than Functional Competence.} Each column compares how the evolution of formal competence (top) and functional competence (bottom) tracks the evolution of brain alignment during training. 
        The $R^2$ values quantify the strength of this relationship, with higher values in formal competence suggesting it as the key driver of the observed brain alignment. \textbf{(a)}: The data averaged across models of five different sizes.
        \textbf{(b-d)}: the same comparison as in (a), but with comparisons were made for models from the Pythia suite with three different sizes. 
    }
    \label{fig:lineplot-correlations}
\end{figure*}

\paragraph{Key Components of Transformers}
To further isolate the key elements responsible for brain alignment in untrained parameter models, we perform an ablation study on the architectural components of \modelname{Transformer-v2} using a single block (Figure \ref{fig:untrained-models}(c)). By focusing on the untrained model, we isolate the effect of architecture alone, without confounding influences from training. The architectural components analyzed are labeled on the left of each bar in Figure \ref{fig:untrained-models}(b).
\code{Attn} refers to all components inside the lower box in Figure \ref{fig:untrained-models}(c), including the first layer norm, multi-head attention, and the residual connection that follows.
\code{MLP} corresponds to the components in the upper box, comprising the post-attention layer norm, MLP, and the subsequent residual layer.
\code{Pos} represents the addition of positional embeddings to token embeddings.
\code{Tokens} means the model directly returns the raw token embeddings without further processing.
This systematic ablation helps pinpoint the components that contribute most to brain alignment.
Once again, we observe that integration across tokens, via attention mechanisms and positional encoding, yields the highest brain alignment.
Further, we found that untrained parameter models perform better than chance-level performance on formal competence benchmarks, mirroring their non-zero brain alignment. In contrast, functional competence benchmarks remain at chance level for untrained models. This further supports the finding that brain alignment is primarily driven by formal, rather than functional, linguistic competence. (see Figure \ref{fig:untrained-models}(d)).

\begin{figure*}
    \centering
    \includegraphics[width=1\linewidth]{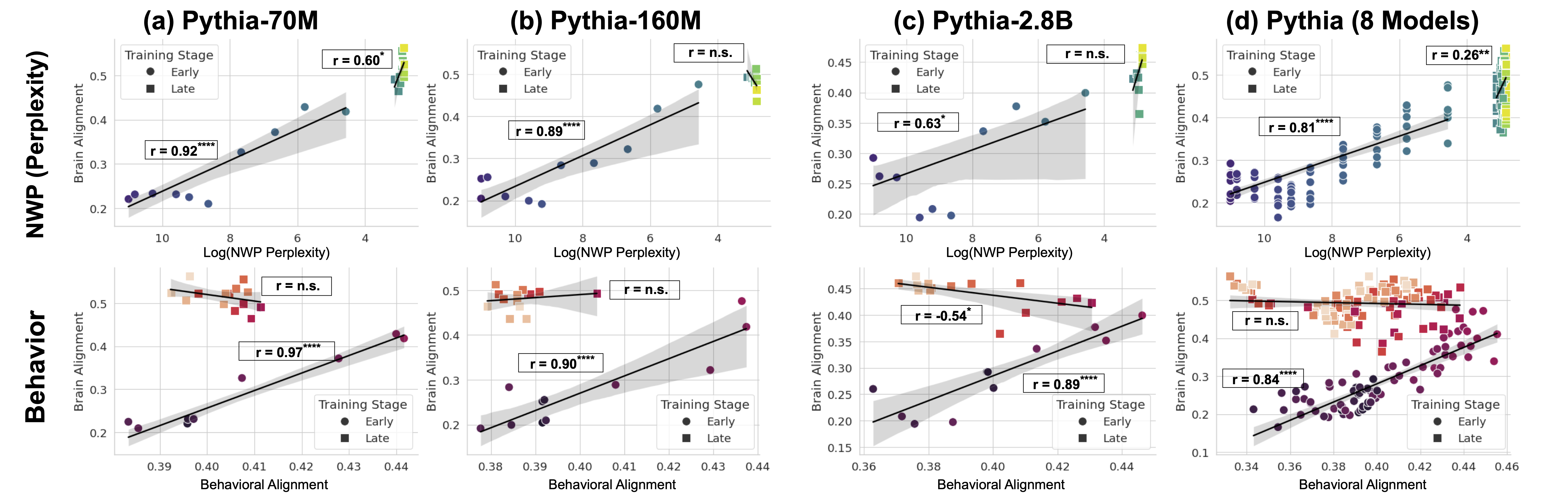}
    \caption{
            \textbf{NWP and Behavioral Alignment Correlate with Brain Alignment Only in Early Training.}  
            \textbf{(Top Row)}: Correlation between brain alignment and language modeling loss shows a strong, significant relationship during early training (up to 2B tokens). While this correlation weakens in later stages (up to \textasciitilde300B tokens). Results are shown for three models and the average of all 8 models (last column). 
            \textbf{(Bottom Row)}: The same analysis, but for the correlation between brain alignment and behavioral alignment, revealing a similar trend—strong correlation early in training, but no significant relationship as models surpass human proficiency.  
    }
    \label{fig:corr-ppl-behavior}
\end{figure*}

\subsection{Brain Alignment Over Training}
\label{sec:alignment-training}

Having established the architectural components that make an untrained model brain-aligned in the previous section, we now investigate how brain alignment evolves during training. To do so, we use the Pythia model suite \cite{pythia}, which consists of models of various sizes, all trained on the same $\sim$300B tokens, with publicly available intermediate checkpoints. We report results for a model from a different family, \modelname{SmolLM2-360M} \citep{allal2025smollm2}, which provides checkpoints at 250B-token intervals, in Appendix \ref{app:smollm2-results}.


Figure \ref{fig:bscore-training} illustrates the brain alignment of six Pythia models across five brain recording datasets at 34 training checkpoints, spanning approximately 300B tokens. Each panel presents checkpoints that are logarithmically spaced up to the vertical line, emphasizing the early-stage increase in brain alignment, which occurs within the first 5.6\% of training time. Beyond this point, the panels display the remaining training period, where brain alignment stabilizes. More specifically, we observe the following trend: 
(1) Brain alignment is similar to the untrained model until approximately 128M tokens.
(2) A sharp increase follows, peaking around 8B tokens.
(3) Brain alignment then saturates for the remainder of training.
Despite the vast difference in model sizes shown in Figure \ref{fig:bscore-training}, the trajectory of brain alignment is remarkably similar. 

\paragraph{Alignment Tracks Formal Competence}
Following the observation that brain alignment plateaus early in training, we next investigate how this relates to the emergence of formal and functional linguistic competence in LLMs. Figure \ref{fig:lineplot-correlations} displays the average brain alignment alongside the average performance on formal competence benchmarks (top row) and functional competence benchmarks (bottom row). This is shown for three Pythia models (1B, 2.8B, and 6.9B parameters) and the average of five Pythia models (first column) across the training process.
To quantify this relationship, we train a ridge regression model (with a single scalar weight) to predict brain alignment scores from benchmark scores using 10-fold cross-validation. The average R-squared value across these folds serves as our metric for comparing the relationship between formal/functional linguistic competence and brain alignment. These R-squared values are shown in each panel of Figure \ref{fig:lineplot-correlations}.
Finally, we perform a Wilcoxon signed-rank test on the distributions of R-squared values. This test reveals that formal linguistic competence is significantly more strongly correlated with brain alignment than functional competence (W = 0.0, p $<$ 0.002). 
One possible explanation for why brain alignment emerges before formal linguistic competence is that existing LLM benchmarks assess performance using discrete accuracy thresholds (hard metrics), rather than capturing the gradual progression of competence through more nuanced, continuous measures (soft metrics) \citep{Schaeffer2023AreEA}. We show the individual benchmark scores across all checkpoints in Figure \ref{fig:competence-finegrained} in Appendix \ref{app:formal-functional-scores}.

\subsection{LLMs Lose Behavioral Alignment}
\label{sec:llm-behavioral-alignment}

Do language models that improve in next-word prediction remain aligned with human behavioral and neural responses, or do they diverge as they surpass human proficiency? To answer this question we use the \datasetname{Futrell2018} benchmark, which has been widely used in previous research to measure linguistic behavior \citep{futrell_natural_2018, schrimpf-pnas, aw2023instructiontuning}. This dataset consists of self-paced reading times for naturalistic story materials from 180 participants. Per-word reading times provide a measure of incremental comprehension difficulty, a cornerstone of psycholinguistic research for testing theories of sentence comprehension \citep{Gibson1998, Smith2013, brothers2021word, shain2024large}. We measure alignment by calculating the Pearson correlation between a model's cross-entropy loss for a specific token in the sequence and the average human per-word reading time. The loss for words that comprise multiple tokens is added together before computing the correlation.

Early in training, LLMs align with this pattern, but as they surpass human proficiency \citep{Shlegeris2022LMNWP}, their perplexity drops and they begin encoding statistical regularities that diverge from human intuition \citep{oh2023does, steuer2023large}. This shift correlates with a decline in behavioral alignment, suggesting that superhuman models rely on different mechanisms than those underlying human language comprehension.
Figure \ref{fig:corr-ppl-behavior} shows that brain alignment initially correlates with perplexity and behavioral alignment, but only during the early stages of training (up to \textasciitilde2B tokens). Beyond this point, these correlations diminish. In larger models, we observe a negative correlation between brain alignment and behavioral alignment in the later stages of training. This trend reinforces that early training aligns LLMs with human-like processing as also observed in earlier stages, while in later stages their language mechanisms diverge from humans.


\section{Conclusion}
In this work, we investigate how brain alignment in LLMs evolves throughout training, revealing different learning processes at play. We demonstrate that alignment with the human language network (LN) primarily correlates with formal linguistic competence \cite{mahowald2024dissociating}, peaking and saturating early in training. In contrast, functional linguistic competence, which involves world knowledge and reasoning, continues to grow beyond this stage. These findings suggest that the LN primarily encodes syntactic and compositional structure, in line with the literature of language neuroscience \cite{Fedorenko2024}, while broader linguistic functions may rely on other cognitive systems beyond the LN. 
This developmental approach reveals when brain-like representations emerge, offering a dynamic perspective compared to prior work focused on fully trained models. For example, \citet{oota2023jointprocessinglinguisticproperties} demonstrated that syntactic structure contributes to alignment by selectively removing specific properties from already trained models. In contrast, we show that formal linguistic competence actively drives brain alignment during the early phases of training. Similarly, \citet{hosseini2024artificial} reported that models achieve strong alignment with limited data; we identify why: the brain-like representations emerge as soon as core formal linguistic knowledge is acquired. Further, their study evaluated only four training checkpoints and 2 models on a single dataset (\datasetname{Pereira2018}). Our study evaluated eight models (14M–6.7B parameters) across 34 checkpoints spanning 300B tokens, and used five neural benchmarks within a rigorous brain‑scoring framework. This extensive design enabled fine‑grained correlations with both formal and functional linguistic benchmarks and ensured our results are robust and generalizable.



We also show that model size is not a reliable predictor of brain alignment when controlling for the number of features (see Appendix \ref{app:model-size}). Instead, alignment is shaped by architectural inductive biases, token integration mechanisms, and training dynamics. 
Our standardized brain-scoring framework eliminates contextualization biases from previous work, ensuring more rigorous evaluations. 
Finally, we demonstrate that current brain alignment benchmarks are not saturated, indicating that LLMs can still be improved in modeling human language processing.
Together, these findings challenge prior assumptions about how alignment emerges in LLMs and provide new insights into the relationship between artificial and biological language processing.

\section*{Limitations}
While this study offers a comprehensive analysis of brain alignment in LLMs, several open questions remain. If functional competence extends beyond the language network, future work should explore which additional brain regions LLMs align with as they develop reasoning and world knowledge, particularly in other cognitive networks like the multiple demand \citep{Duncan2000} or theory of mind network \citep{Saxe2003, saxe2006}. Our findings suggest that LLM brain alignment studies should be broadened from the LN to downstream representations underlying other parts of cognition. This raises the question of whether specific transformer units specialize in formal vs. functional linguistic competence \citep{AlKhamissi2024TheLN}.

One other limitation of our study is that we rely exclusively on brain data collected from experiments conducted with English stimuli. As such, we do not explore whether our findings generalize across languages. This remains an open question and warrants further investigation. That said, evidence from cross-linguistic neuroscience research studying 45 languages from 12 language families \citep{MalikMoraleda2022} suggests the existence of a universal language network in the brain that is robust across languages and language families, both in topography and core functional properties.

Finally, a key question remains: Does LLM alignment evolution mirror human language acquisition? Comparing LLM representations to developmental data could reveal insights into learning trajectories and help differentiate formal from functional language learning. Expanding brain-scoring benchmarks and incorporating multimodal models will help address these questions, further bridging the gap between artificial and biological intelligence and deepening our understanding of how both systems process and represent language.

\section*{Ethical Statement}
This research relies on previously published neuroimaging (fMRI, ECoG) and behavioral datasets, collected by the original research groups under their institutional ethical guidelines with informed consent and IRB/ethics approval. Our work involved only secondary analysis of de-identified data, with no new data collection or direct participant interaction, and we remain committed to using such data responsibly and respectfully.

\section*{Acknowledgments}
We thank the members of the EPFL NeuroAI and NLP labs for their valuable feedback and insightful suggestions. We also gratefully acknowledge the support of the Swiss National Science Foundation (No. 215390), Innosuisse (PFFS-21-29), the EPFL Center for Imaging, Sony Group Corporation, and a Meta LLM Evaluation Research Grant.


\bibliography{references}

\clearpage

\appendix

\begin{table*}[ht]
    \centering
    \setlength{\tabcolsep}{2pt}
    \begin{tabular}{lllll}
    \toprule 
    \textbf{Dataset} & \textbf{Modality} & \textbf{Presentation} & \textbf{Stimulus Example} \\
    \midrule
    \textsc{Pereira2018} & fMRI & Reading & Accordions produce sound with bellows ... \\
    \textsc{Blank2014} & fMRI & Listening & A clear and joyous day it was and out on the wide ...   \\
    \textsc{Fedorenko2016} & ECoG & Reading & `ALEX', `WAS', `TIRED', `SO', `HE', `TOOK', ... \\
    \textsc{Tuckute2024} & fMRI & Reading  & The judge spoke, breaking the silence. \\
    \textsc{Narratives} & fMRI & Listening  & Okay so getting back to our story about uh Lucy ... \\
    \midrule
    \textsc{Futrell2018} & Reading Times & Reading & A clear and joyous day it was and out on the wide ... \\
    \bottomrule \\
    \end{tabular}
    \caption{
    \textbf{Datasets Used for Evaluating Model Alignment.} Neuroimaging datasets were collected via either functional magnetic resonance imaging (fMRI) or electrocorticography (ECoG). Stimuli range from short sentences (\datasetname{Fedorenko2016}, \datasetname{Tuckute2024}) to paragraphs (\datasetname{Pereira2018}) and entire stories (\datasetname{Blank2014}, \datasetname{Narratives}, \datasetname{Futrell2018}) and were presented either visually or auditorily. \datasetname{Futrell2018} is a behavioral dataset.}
    \label{tab:dataset}
\end{table*}

\begin{figure*}[t]
    \centering
    \includegraphics[width=1\linewidth]{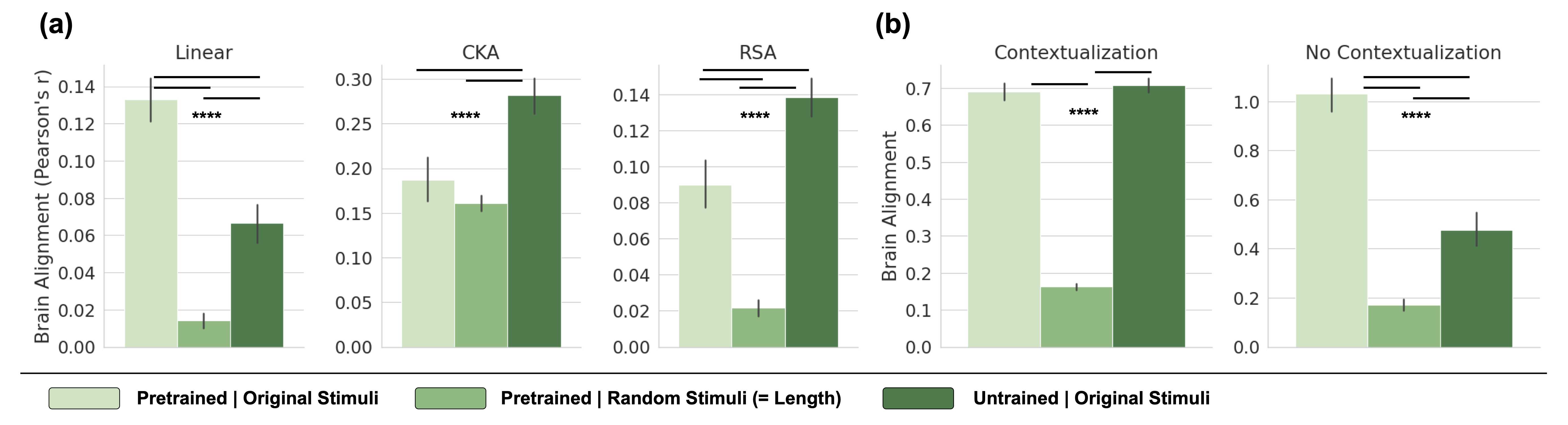}
    \caption{
        \textbf{Evaluating Brain Alignment with Linear Predictivity and No Contextualization is Most Stringent}.
        \textbf{(a)} Average brain alignment across 8 Pythia models under three conditions: (1) a pretrained model processing the original stimuli, (2) a pretrained model processing \emph{random} sequences of the same length (averaged over five random seeds) as a control condition, and (3) the model with untrained parameters processing the original stimuli. The linear predictivity metric differentiates between meaningful and random stimuli most strongly, while RSA and CKA overestimate alignment. 
        \textbf{(b)} Brain alignment on the \datasetname{Pereira2018} dataset under two cross-validation schemes: with contextualization (random sentence split) and without contextualization (story-based split). 
    }
    \label{fig:metrics-context}
\end{figure*}

\section*{Appendix}

\section{Neuroimaging \& Behavioral Datasets}
\label{app:datasets}

Table \ref{tab:dataset} shows the different neuroimaging and behavioral datasets used in this work, along with the dataset modality, presentation mode, and a stimulus example.

\subsection{Neuroimaging Datasets}
\label{app:neural-datasets}

\paragraph{\cite{pereira_toward_2018}} This dataset consists of fMRI activations (blood-oxygen-level-dependent; BOLD responses) recorded as participants read short passages presented one sentence at a time for 4 s. The dataset is composed of two distinct experiments: one with 9 subjects presented with 384 sentences, and another with 6 subjects presented with 243 sentences each. The passages in each experiment spanned 24 different topics. The results reported for this dataset are the average alignment across both experiments after normalizing with their respective cross-subject consistency estimates.

\paragraph{\cite{blank_functional_2014}} This dataset also involves fMRI signals but recorded from only 12 functional regions of interest (fROI) instead of the higher resolution signal used by \citet{pereira_toward_2018}. The data was collected from 5 participants as they listened to 8 long naturalistic stories that were adapted from existing fairy tales and short stories \citep{futrell_natural_2018}. Each story was approximately 5 minutes long, averaging up to 165 sentences, providing a much longer context length than the other neuroimaging datasets. When measuring brain alignment, we use the input stimuli of the last 32 TRs as the model's context.

\paragraph{\cite{fedorenko2016}} This dataset captures ECoG signals from 5 participants as they read 8-word-long sentences presented one word at a time for 450 or 700 ms. Following \cite{schrimpf-pnas} we select the 52/80 sentences that were presented to all participants. 

\paragraph{\cite{tuckute2024driving}} 
In this dataset, 5 participants read 1000 6-word sentences presented one sentence at a time for 2 s. BOLD responses from voxels in the language network were averaged within each participant and then across participants to yield an overall average language network response to each sentence. The stimuli used span a large part of the linguistic space, enabling model-brain comparisons across a wide range of single sentences. Sentence presentation order was randomized across participants. In combination with the diversity in linguistic materials, this dataset presents a particularly challenging dataset for model evaluation.

\paragraph{Narratives Dataset \citep{narratives}}
This dataset consists of fMRI data collected while human subjects listened to 27 diverse spoken story stimuli. The collection includes 345 subjects, 891 functional scans, and approximately 4.6 hours of unique audio stimuli. For our story-based analysis, we focused on 5 participants who each listened to both the \datasetname{Lucy} and \datasetname{Tunnel} stories. Since functional localization was not performed in the \datasetname{Narratives} dataset, we approximated language regions by extracting the top-10\% voxels from each anatomically defined language region according to a probabilistic atlas for the human language system \citep{Lipkin2022}. Due to the limited corpus of two stories, traditional 10-fold cross-validation was not feasible. To implement topic-based splitting while maintaining methodological rigor, we partitioned each story into $n$ distinct segments, with each segment functioning as an independent narrative unit. This segmentation approach effectively prevented cross-contamination of contextual information between splits, thereby preserving the integrity of our evaluation framework.

\subsection{Behavioral Dataset}

\paragraph{\citep{futrell_natural_2018}}
This dataset consists of self-paced reading times for each word from 180
participants. The stimuli include 10 stories from the Natural Stories Corpus \citep{futrell_natural_2018},
similar to \datasetname{Blank2014}. Each participant read between 5 and all 10 stories.

\section{Rigorous Brain-Scoring}
\label{app:rigorous-brain-scoring}
Despite progress in linking LLMs to neural activity, there’s no standard for comparing brain alignment across datasets and conditions. Here, we aim to establish a set of desiderata for evaluating brain alignment. For a model to be considered truly brain-aligned, two key criteria must be met. First, high alignment scores should indicate that the model captures stimulus-driven responses---meaning that when presented with a random sequence of tokens, alignment should drop significantly compared to original linguistic stimuli. Second, a brain-aligned model should generalize effectively to new linguistic contexts rather than overfitting to specific examples. We address these two points in Section \ref{sec:rigorous-brainscoring} to justify our choice of metric and cross-validation scheme for each dataset (see Figure \ref{fig:metrics-context}). For all benchmarks, we localize language-selective units, which is consistent with neural site selection in neuroscience experiments and allows for fair comparisons across models irrespective of model size \cite{AlKhamissi2024TheLN}. A key limitation of previous methods is their reliance on the raw hidden state dimensions, which inherently favors larger models by providing a greater feature space and artificially inflating alignment scores.

\begin{table*}[ht]
\centering
\resizebox{\textwidth}{!}{%
\begin{tabular}{l|ccccc|c}
\toprule
\textbf{Num Tokens} & \textbf{Pereira2018} & \textbf{Blank2014} & \textbf{Tuckute2024} & \textbf{Fedorenko2016} & \textbf{Narratives} & \textbf{Avg} \\
\midrule
250B  & 1.00 & 0.19 & 0.47 & 0.78 & 0.04 & 0.50 \\
500B  & 0.97 & 0.08 & 0.51 & 0.87 & 0.04 & 0.49 \\
750B  & 0.99 & 0.08 & 0.52 & 0.78 & 0.04 & 0.48 \\
1T    & 1.07 & 0.12 & 0.55 & 0.84 & 0.04 & 0.52 \\
1.25T & 1.00 & 0.12 & 0.50 & 0.82 & 0.03 & 0.49 \\
1.5T  & 1.00 & 0.12 & 0.52 & 0.79 & 0.03 & 0.49 \\
1.75T & 0.96 & 0.13 & 0.48 & 0.79 & 0.04 & 0.48 \\
2T    & 1.05 & 0.15 & 0.56 & 0.84 & 0.04 & 0.53 \\
2.25T & 1.08 & 0.16 & 0.55 & 0.75 & 0.04 & 0.51 \\
2.5T  & 1.12 & 0.17 & 0.52 & 0.72 & 0.01 & 0.51 \\
2.75T & 1.13 & 0.12 & 0.49 & 0.75 & 0.04 & 0.49 \\
3T    & 1.03 & 0.26 & 0.51 & 0.55 & 0.01 & 0.47 \\
3.25T & 1.02 & 0.13 & 0.52 & 0.68 & 0.02 & 0.47 \\
3.5T  & 1.04 & 0.14 & 0.52 & 0.72 & 0.04 & 0.49 \\
3.75T & 1.14 & 0.06 & 0.57 & 0.84 & 0.03 & 0.53 \\
4T    & 1.05 & 0.13 & 0.63 & 0.82 & 0.05 & 0.54 \\
\bottomrule
\end{tabular}}
\caption{
    \textbf{Brain Alignment Performance of \modelname{SmolLM2-360M} Across Training Checkpoints.} Reported scores correspond to normalized correlations with neural responses from five benchmark datasets (Pereira2018, Blank2014, Tuckute2024, Fedorenko2016, Narratives), along with their average (Avg). These results assess the extent to which the model’s internal representations align with activity in the human language network.
}
\label{tab:smollm2-brain}
\end{table*}

\section{Brain-Score Using Additional Metrics}
\label{app:add-metrics}


\paragraph{Centered Kernel Alignment (CKA)} \citet{cka} introduced CKA as a substitute for Canonical Correlation Analysis (CCA) to assess the similarity between neural network representations. Unlike linear predictivity, it is a non-parameteric metric and therefore does not require any additional training. CKA is particularly effective with high-dimensional representations, and its reliability in identifying correspondences between representations in networks trained from different initializations \citep{cka}.
  
\paragraph{Representational Similarity Analysis (RSA)} 
\cite{rdm} introduced RDMs as a solution to the challenge of integrating brain-activity measurements, behavioral observations, and computational models in systems neuroscience. RDMs are part of a broader analytical framework referred to as representational similarity analysis (RSA). In practical terms, to compute the dissimilarity matrix for an $N$-dimensional network's responses to $M$ different stimuli, an $M$×$M$ matrix of distances between all pairs of evoked responses is generated for both brain activity and the language model's activations \cite{harvey2023duality}. The correlation between these two matrices is then used as a measure of brain alignment.

\begin{table*}[ht]
\centering
\resizebox{\textwidth}{!}{%
\begin{tabular}{l|ccccccccc|c}
\toprule
\textbf{Num Tokens} & \textbf{BLiMP} & \textbf{SyntaxGym} & \textbf{Avg (Formal)} & \textbf{ARC-Easy} & \textbf{ARC-Challenge} & \textbf{Social-IQA} & \textbf{PIQA} & \textbf{WinoGrande} & \textbf{HellaSwag} & \textbf{Avg (Functional)} \\
\midrule
250B  & 0.81 & 0.80 & 0.81 & 0.33 & 0.66 & 0.35 & 0.70 & 0.55 & 0.47 & 0.52 \\
500B  & 0.80 & 0.78 & 0.79 & 0.78 & 0.66 & 0.35 & 0.70 & 0.56 & 0.49 & 0.53 \\
750B  & 0.80 & 0.82 & 0.81 & 0.69 & 0.69 & 0.34 & 0.71 & 0.57 & 0.50 & 0.53 \\
1T    & 0.81 & 0.78 & 0.80 & 0.69 & 0.69 & 0.35 & 0.71 & 0.57 & 0.50 & 0.54 \\
1.25T & 0.81 & 0.78 & 0.79 & 0.68 & 0.68 & 0.35 & 0.71 & 0.57 & 0.51 & 0.54 \\
1.5T  & 0.81 & 0.80 & 0.80 & 0.69 & 0.68 & 0.35 & 0.72 & 0.56 & 0.51 & 0.54 \\
1.75T & 0.80 & 0.79 & 0.79 & 0.68 & 0.68 & 0.36 & 0.72 & 0.59 & 0.51 & 0.54 \\
2T    & 0.81 & 0.81 & 0.81 & 0.69 & 0.69 & 0.35 & 0.72 & 0.59 & 0.52 & 0.54 \\
2.25T & 0.81 & 0.82 & 0.81 & 0.68 & 0.68 & 0.35 & 0.71 & 0.59 & 0.51 & 0.54 \\
2.5T  & 0.81 & 0.82 & 0.82 & 0.68 & 0.68 & 0.36 & 0.70 & 0.56 & 0.52 & 0.54 \\
2.75T & 0.81 & 0.82 & 0.81 & 0.25 & 0.23 & 0.35 & 0.50 & 0.57 & 0.50 & 0.50 \\
3T    & 0.81 & 0.81 & 0.81 & 0.25 & 0.23 & 0.35 & 0.50 & 0.57 & 0.50 & 0.50 \\
3.25T & 0.81 & 0.77 & 0.79 & 0.67 & 0.67 & 0.34 & 0.67 & 0.57 & 0.51 & 0.52 \\
3.5T  & 0.81 & 0.79 & 0.80 & 0.71 & 0.71 & 0.38 & 0.72 & 0.58 & 0.53 & 0.55 \\
3.75T & 0.80 & 0.78 & 0.79 & 0.72 & 0.72 & 0.58 & 0.58 & 0.54 & 0.56 & 0.56 \\
4T    & 0.81 & 0.79 & 0.80 & 0.73 & 0.73 & 0.39 & 0.74 & 0.61 & 0.56 & 0.57 \\
\bottomrule
\end{tabular}}
\caption{
    \textbf{Performance of \modelname{SmolLM2-360M} on Formal and Functional Linguistic Benchmarks Across Training Checkpoints.} Formal competence is measured using BLiMP and SyntaxGym (with averages reported as Avg Formal). Functional competence is measured using ARC-Easy, ARC-Challenge, Social-IQA, PIQA, WinoGrande, and HellaSwag (with averages reported as Avg Functional). Together, these results characterize the relationship between training progression and the development of different aspects of linguistic ability.
}
\label{tab:smollm2-formal-functional}
\end{table*}

\section{Brain Alignment Over Training}
\label{app:brain-alignment-training}

\begin{figure*}[t]
    \centering
    \includegraphics[width=1\linewidth]{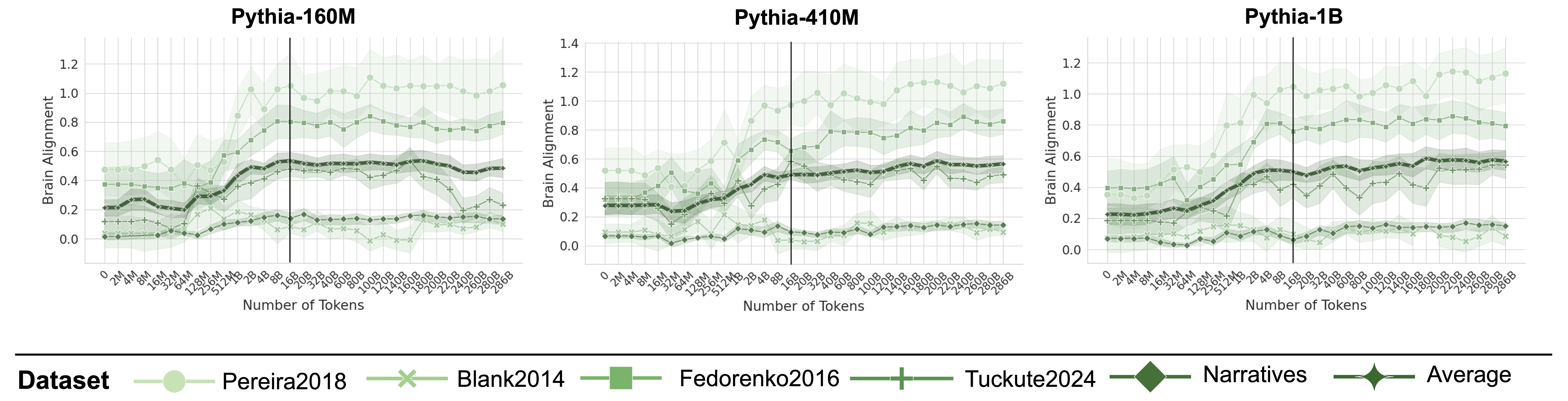}
    \caption{
        \textbf{Brain Alignment Saturates Early on in Training.} 
        Plots complementing Figure \ref{fig:bscore-training} showing the brain alignment scores of three other models from the Pythia model suite with varying sizes (log x-axis up to 16B tokens, uneven spacing after black line). Scores are normalized by their cross-subject consistency scores. Alignment quickly peaks around 2–8B tokens before saturating or declining, regardless of model size.
    }
    \label{fig:bscore-training-appendix}
\end{figure*}

Figure \ref{fig:bscore-training-appendix} complements Figure \ref{fig:bscore-training} in the main paper, illustrating that brain alignment saturates early on in training for all models analyzed in this work. 

\section{Formal \& Functional Scores}
\label{app:formal-functional-scores}

\begin{figure*}[ht]
    \centering
    \includegraphics[width=1\linewidth]{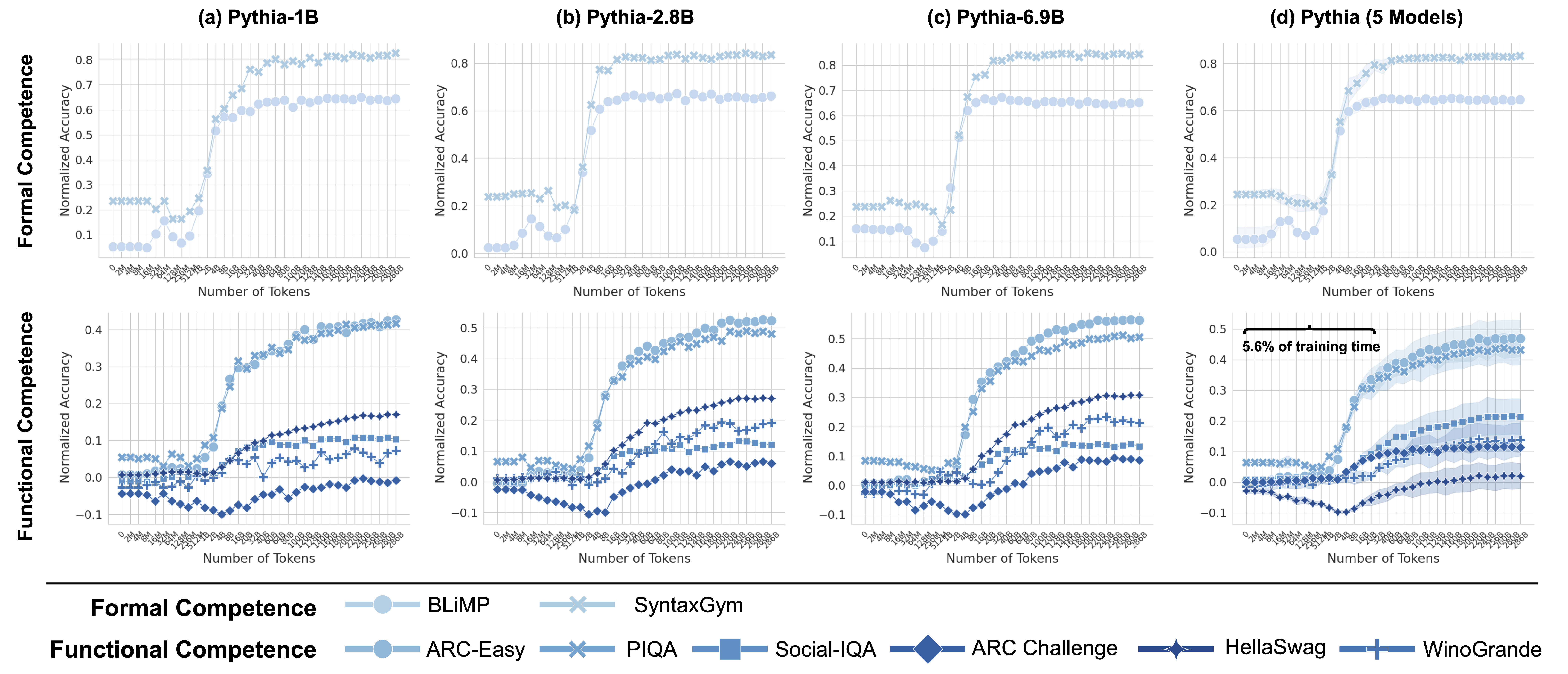}
    \caption{
    \textbf{Individual Benchmark Scores for Formal and Functional Competence.} \textbf{(a-c)}: each column shows the evolution of individual benchmark scores for formal competence (top) and functional competence (bottom) during training. Data is presented for Pythia models of three different sizes. \textbf{(d)}: the same as (a–c), with data averaged across models of five different sizes.
    }
    \label{fig:competence-finegrained}
\end{figure*}

Figure \ref{fig:competence-finegrained} presents the individual benchmark scores for both formal and functional linguistic competence across training. Formal benchmarks peak early, mirroring the trajectory of brain alignment, and remain saturated throughout training. In contrast, functional benchmarks continue to improve, reflecting the models’ increasing ability to acquire factual knowledge and reasoning skills as they are trained on significantly more tokens using next-word prediction.

\section{Results on \modelname{SmolLM2-360M}}
\label{app:smollm2-results}

To assess the generalizability of our findings, we replicated our experiments using a model from a different language family. Specifically, we evaluated multiple training checkpoints of \modelname{SmolLM2-360M} on the brain alignment, formal, and functional linguistic competence benchmarks. Since SmolLM2 only provides checkpoints at intervals of 250B tokens, we cannot capture the gradual emergence of brain alignment and formal competence, both of which typically saturate around 4B–8B tokens. Given this limitation, our hypothesis was that brain alignment and formal competence would remain largely stable across these checkpoints, while functional competence would continue to improve. The results are consistent with this hypothesis as shown in Tables \ref{tab:smollm2-brain} and \ref{tab:smollm2-formal-functional}.

\section{Role of Weight Initialization}

\begin{figure}[h]
    \centering
    \includegraphics[width=1\linewidth]{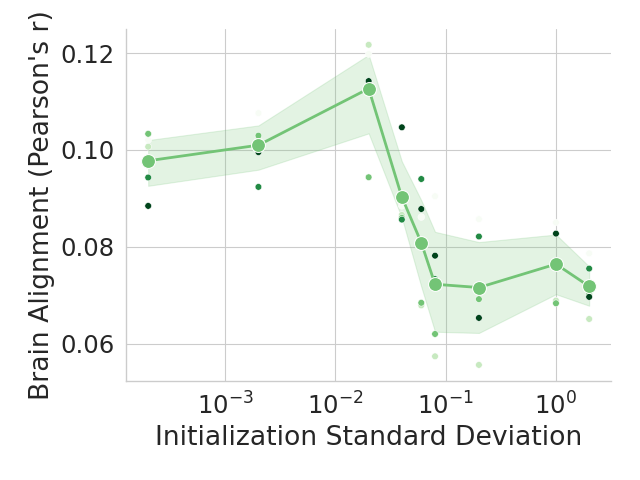}
    \caption{
        \textbf{Role of Weight Initialization on Brain Alignment in Untrained Models}
        The default initialization standard deviation in the HuggingFace library (sd = 0.02) yields the highest brain alignment for untrained models, suggesting that initialization choices play a crucial role in shaping alignment even before training begins.
    }
    \label{fig:weight-init}
\end{figure}

Figure \ref{fig:weight-init} examines the effect of weight initialization variance on brain alignment in untrained models. We systematically vary the initialization standard deviation (sd) and find that the default HuggingFace \cite{Wolf2019HuggingFacesTS}  initialization (sd = 0.02) achieves the highest alignment across datasets. This suggests that even before training begins, the choice of initialization can significantly influence how well a model’s representations align with neural activity. 
This finding raises an intriguing hypothesis: could brain alignment, a computationally inexpensive metric, serve as a useful heuristic for selecting optimal initialization parameters? If so, it could help models learn tasks more efficiently and converge faster, reducing the need for extensive trial-and-error in training from scratch.
The results highlight the importance of architectural inductive biases and suggest that brain alignment may serve as a useful heuristic for optimizing model initialization.

\section{Effect of Number of Units on Brain Alignment}
\label{app:number-of-units}

\begin{figure}[ht]
    \centering
    \includegraphics[width=1\linewidth]{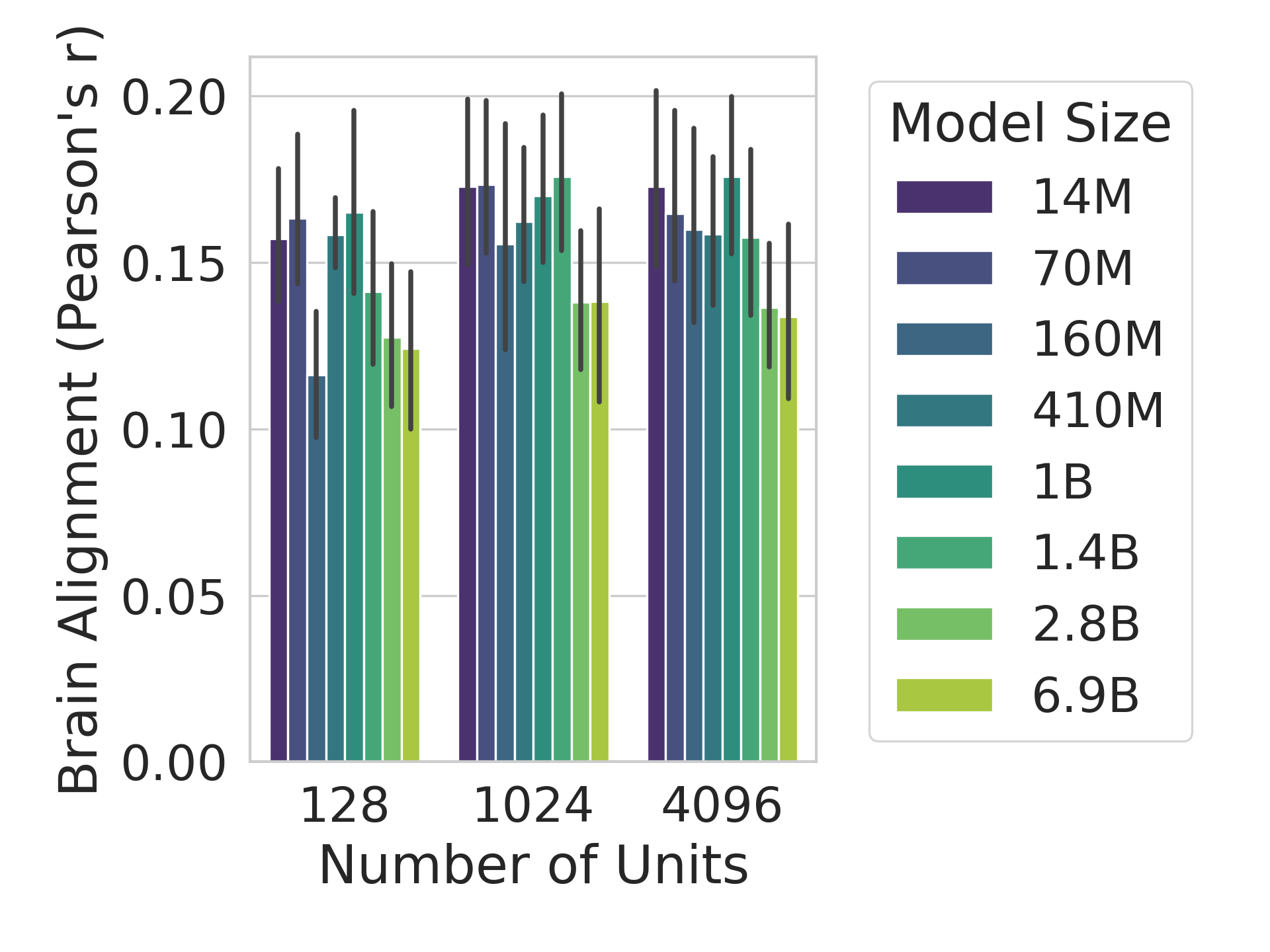}
    \caption{
        \textbf{The Effect of the Number of Localized Units on Final Brain Alignment}
        Brain alignment is evaluated after localizing 128, 1024, and 4096 units. While increasing the number of units slightly affects overall alignment, the relative ranking of models remains largely unchanged, indicating that model comparisons are robust to the choice of unit count.
    }
    \label{fig:num-units-effect}
\end{figure}

Figure \ref{fig:num-units-effect} illustrates the impact of localizing more units on final brain alignment across the eight Pythia models used in this study. We find that increasing the number of units has minimal impact on the relative ranking of models, with only a slight increase in average alignment. Additionally, model size does not influence brain alignment once the number of units is controlled, reinforcing the idea that alignment is driven by feature selection rather than scale.

\begin{figure*}[ht!]
    \centering
    \includegraphics[width=1\linewidth]{figures/brain-score-llms-brain-alignment-v1.drawio.png}
    \caption{
        \textbf{Brain Alignment with the Language Network vs. V1 Across Training.}
        Raw brain alignment scores (Pearson's r) of three Pythia models of varying sizes are shown on the \datasetname{Pereira2018} dataset. The x-axis (log-scaled up to 16B tokens; then evenly spaced after the black line every 20B tokens) represents training progress. Alignment with V1, an early visual region, remains stable throughout training, while alignment with the language network (LN) increases around 4B tokens before plateauing.
    }

    \label{fig:brain-alignment-v1}
\end{figure*}

\section{Model Size Does Not Predict Alignment}
\label{app:model-size}

\begin{figure}[t]
    \centering
    \includegraphics[width=1\linewidth]{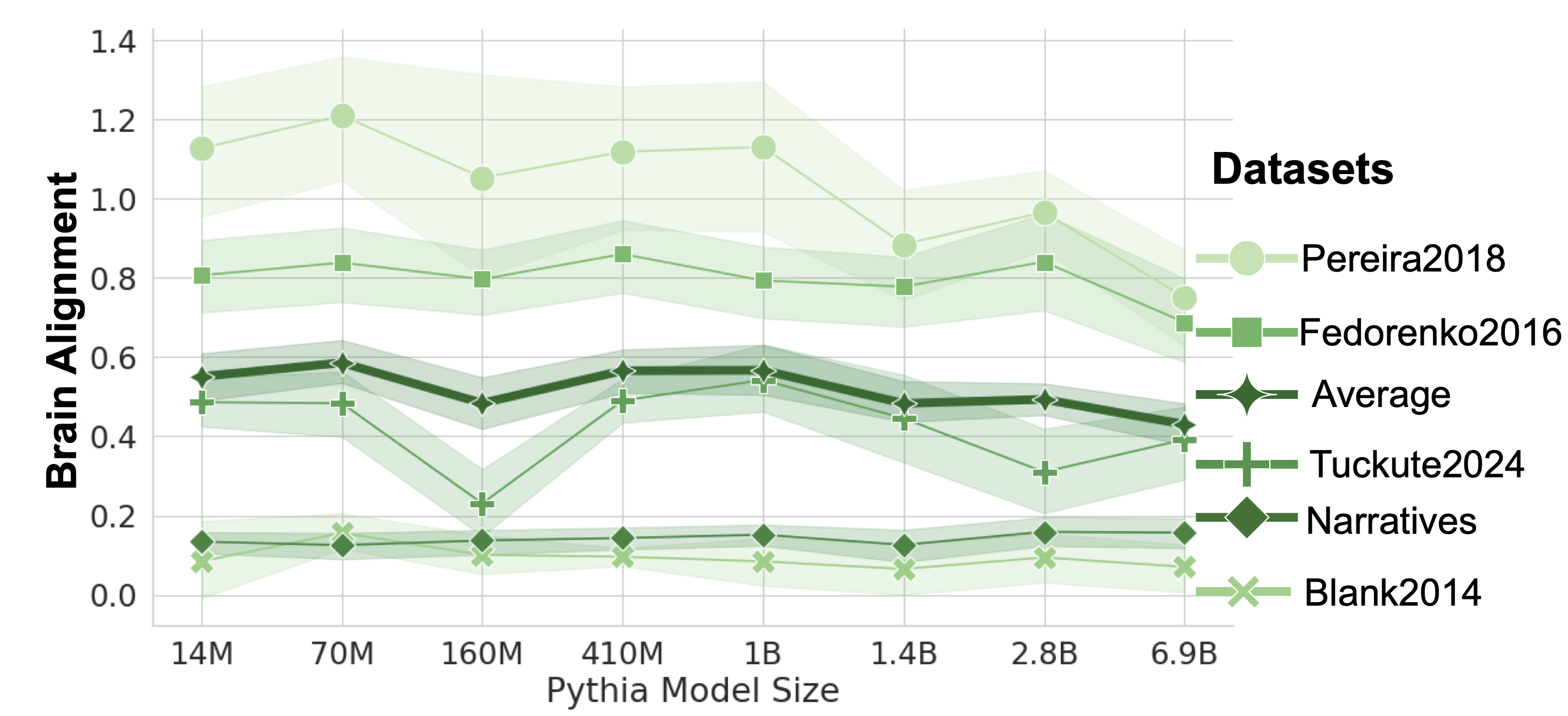}
    \caption{
        \textbf{Model Size Does Not Predict Brain Alignment when localizing a fixed set of language units}.
        Brain alignment across model sizes in the Pythia suite, measured at their final training checkpoints. Brain alignment is shown for each dataset, along with the average score across datasets, for eight models of varying sizes.
    }
    \label{fig:model-size}
\end{figure}

Figure \ref{fig:model-size} presents the brain alignment for each dataset, along with the average alignment across datasets, for eight models of varying sizes from the Pythia model suite (final checkpoint). Contrary to the assumption that larger models exhibit higher brain alignment \cite{aw2023instructiontuning}, we observe a decline in average alignment starting from 1B parameters up to 6.9B parameters, when controlling for feature size. This analysis is made possible by functional localization, which allows us to extract a fixed number of units from each model, rather than relying on hidden state dimensions, as done in previous studies. This approach ensures a fairer comparison among models. We show in Appendix \ref{app:number-of-units} that increasing the number of localized units has minimal impact on the relative ranking of the models. Additionally, these findings align with expectations in the neuroscience language community, where it is widely believed that human language processing does not require superhuman-scale models to capture neural activity in the brain’s language network.

\section{Alignment with Other Brain Regions}
\label{app:other-brain-regions}




As a control, we also examine alignment with non-language brain regions. Specifically, Figure~\ref{fig:brain-alignment-v1} shows the brain alignment of three Pythia models with both the language network (LN) and V1---an early visual cortex region---on the \datasetname{Pereira2018} dataset. While alignment with the LN increases early in training (around 4B tokens) and then saturates, alignment with V1 remains largely unchanged throughout training. This divergence highlights a key aspect of LLM representations: they do not appear to encode low-level perceptual features, such as those processed in early visual areas. If models were learning perceptual structure from the stimuli, we would expect alignment with V1 to increase alongside LN alignment. Instead, the stability of V1 alignment across training suggests that language models selectively develop internal representations that align with higher-order linguistic processing rather than general sensory processing.


One reason for not measuring alignment against other higher-level cognitive brain regions such as the default mode network (DMN), the multiple demand network (MD) or the theory of mind network (ToM) is due to a major limitation in current neuroimaging datasets: the linguistic stimuli used in studies with publicly available datasets (e.g., \datasetname{Pereira2018}) do not reliably engage these higher-level cognitive regions, leading to substantial variability across individuals and thus much lower cross-subject consistency scores. Simply “looking” for alignment in the DMN or MD is therefore insufficient. Instead, we need new datasets that deliberately activate non‑language networks and record item‑level neural responses. For example, most MD studies rely on blocked fMRI designs (e.g., hard vs. easy math), yielding one activation estimate per condition rather than per stimulus. Such coarse measurements limit their utility to evaluate model‑to‑brain correspondence at the granularity of individual items. We expect alignment with the MD network, a brain region involved in logical reasoning, to track functional linguistic competence more than formal competence as models improve on relevant benchmarks. We leave this investigation for future work, pending the availability of suitable datasets.

\section{Cross-Subject Consistency Scores}
\label{app:consistency}

\begin{table}[ht]
    \centering
    \begin{tabular}{lc}
        \toprule 
        \textbf{Benchmark} & \textbf{Consistency Score} \\
        \midrule
         \datasetname{Pereira2018} (Exp 2)$^*$ & 0.086 \\
         \datasetname{Pereira2018} (Exp 3) & 0.144 \\
         \datasetname{Blank2014} &  0.178 \\
         \datasetname{Fedorenko2016} &  0.222 \\
         \datasetname{Tucktue2024} & 0.559  \\
         \datasetname{Narratives} &  0.181 \\
         \midrule 
         \datasetname{Futrell2018} &  0.858 \\
         \bottomrule
    \end{tabular}
    \caption{
        \textbf{Cross-Subject Consistency Scores}
        The values used to normalize the raw Pearson correlation. $^*$\datasetname{Pereira2018} (Exp 2) was computed without extrapolation.
    }
    \label{tab:consistency-scores}
\end{table}

Table \ref{tab:consistency-scores} shows the cross-subject consistency scores computed with extrapolation for the different benchmarks used in this work.

\end{document}